\documentclass{article}
\usepackage[preprint]{colm2026_conference}

\usepackage{microtype}
\usepackage{hyperref}
\usepackage{url}
\usepackage{booktabs}
\usepackage{lineno}

\usepackage{xcolor}
\usepackage{graphicx}

\usepackage{tabularx}
\usepackage{array}

\usepackage{tikz}
\usetikzlibrary{shapes, shadows, positioning, arrows.meta}

\definecolor{darkblue}{rgb}{0, 0, 0.5}
\hypersetup{
    colorlinks=true,
    citecolor=darkblue,
    linkcolor=darkblue,
    urlcolor=darkblue
}

\title{Magic, Madness, Heaven, Sin: 
\\ LLM Output Diversity is Everything, Everywhere, All at Once}

\author{Harnoor Dhingra\thanks{Independent research. For correspondence: \texttt{hadhingra@microsoft.com}} \\
Microsoft \\
}

\begin{document}

\ifcolmsubmission
\linenumbers
\fi

\maketitle

\begin{abstract}
\label{sec:abstract}
Research on Large Language Models (LLMs) studies output variation across generation, reasoning, alignment, and representational analysis, often under the umbrella of \emph{diversity}. Yet the terminology remains fragmented, largely because the normative objectives underlying tasks are rarely made explicit. We introduce the Magic, Madness, Heaven, Sin framework, which models output variation along a homogeneity-heterogeneity axis, where valuation is determined by the task and its normative objective. We organize tasks into four normative contexts: epistemic (factuality), interactional (user utility), societal (representation), and safety (robustness). For each, we examine the failure modes and vocabulary such as hallucination, mode collapse, bias, and erasure through which variation is studied. We apply the framework to analyze all pairwise cross-contextual interactions, revealing that optimizing for one objective, such as improving safety, can inadvertently harm demographic representation or creative diversity. We argue for context-aware evaluation of output variation, reframing it as a property shaped by task objectives rather than a model's intrinsic trait.

\end{abstract}
\section{Introduction}


Large Language Models (LLMs) are expected to produce strictly factual responses to questions such as ``Who is the CEO of Microsoft?'', generate imaginative content in brainstorming tasks, and provide personalized outputs that align with social norms without reinforcing stereotypes. While this ability to modulate output behavior is central to their utility, the scientific vocabulary used to describe these output variations remains deeply fragmented. 

Model output behavior has been studied across a broad range of areas, including natural language generation, question answering, reasoning, alignment, and representational analysis, often under the umbrella of ``diversity'' \cite{guo2025benchmarkinglinguisticdiversitylarge, Murthy_2025, kirk2024understandingeffectsrlhfllm, gallegos-etal-2024-bias, jiang2025artificialhivemindopenendedhomogeneity, lahoti-etal-2023-improving}. Yet this work has largely proceeded within separate task-specific settings. While researchers have identified specific trade-offs between these behaviors, such as the effect of alignment on diversity \citet{kirk2024understandingeffectsrlhfllm, Murthy_2025} or the tradeoff between personalization and stereotyping \cite{kantharuban-etal-2025-stereotype}, these efforts remain largely siloed. 

In this paper, we argue that these seemingly distinct behaviors across tasks can be analyzed through a common lens: variation in model outputs. We model this variation along a continuous axis of \emph{homogeneity} to \emph{heterogeneity}, where responses range from highly consistent and convergent to varied and divergent. Importantly, this variation is not inherently desirable or undesirable—its value depends entirely on the normative objectives of the given task.

Hence, the valuation of output variation — whether homogeneity or heterogeneity is preferred — is determined by the task and its normative objective. As many tasks share similar objectives and valuations, we group them into four normative contexts based on their dominant valuation. We use ``normative context'' (signifies its dominant objective) and ``normative objective'' interchangeably throughout the paper.


\begin{figure}[h]
\centering

\usetikzlibrary{arrows.meta}


\definecolor{magicColor}{HTML}{E3F2FD}  
\definecolor{madnessColor}{HTML}{FFEBEE} 
\definecolor{heavenColor}{HTML}{E8F5E9}  
\definecolor{sinColor}{HTML}{FFF3E0}     

\definecolor{darktext}{HTML}{263238}    
\definecolor{axisColor}{HTML}{222222}
\definecolor{textColor}{HTML}{111111}

\begin{tikzpicture}[
    x=1cm, y=1cm,
    axis/.style={->, >=Stealth, line width=1pt, draw=axisColor},
    card/.style={
        rounded corners=3pt,
        minimum width=3.6cm,
        minimum height=2.0cm,
        inner sep=4pt,
        align=center,
        font=\sffamily\scriptsize,
        text=textColor
    },
    axislabel/.style={
        font=\sffamily\bfseries\tiny,
        text=axisColor
    }
]

\draw[axis] (-3.9,0) -- (3.9,0);
\draw[axis] (0,-2.3) -- (0,2.3);

\node[axislabel] at (0,2.5) {REWARDED};
\node[axislabel] at (0,-2.5) {PENALIZED};
\node[axislabel, anchor=east] at (-4.05,0.15) {HOMOGENEITY};
\node[axislabel, anchor=west] at (4.05,0.15) {HETEROGENEITY};

\node[card, fill=heavenColor] at (-1.9, 1.1) {
\textbf{\scriptsize SAFETY HEAVEN}\\[-1pt]
\textit{\tiny Objective: Robustness}\\
\tiny Homogeneity $\rightarrow$ Compliance
};

\node[card, fill=magicColor] at (1.9, 1.1) {
\textbf{\scriptsize INTERACTIONAL MAGIC}\\[-1pt]
\textit{\tiny Objective: Engagement \& Utility}\\
\tiny Heterogeneity $\rightarrow$ Creativity
};

\node[card, fill=sinColor] at (-1.9, -1.1) {
\textbf{\scriptsize SOCIETAL SIN}\\[-1pt]
\textit{\tiny Objective: Representation}\\
\tiny Homogeneity $\rightarrow$ Erasure
};

\node[card, fill=madnessColor] at (1.9, -1.1) {
\textbf{\scriptsize EPISTEMIC MADNESS}\\[-1pt]
\textit{\tiny Objective: Factuality}\\
\tiny Heterogeneity $\rightarrow$ Hallucination
};

\end{tikzpicture}

\caption{\textbf{The Magic, Madness, Heaven, Sin Framework.} Output variation in LLMs lies on a homogeneity–heterogeneity axis. The valuation of this variation — whether it is rewarded or penalized — is determined by the task and its normative objective. We organize tasks into four normative contexts based on their dominant valuation: heterogeneity enables creativity in interactional settings (\emph{Magic}) but leads to hallucination in epistemic settings (\emph{Madness}), while homogeneity supports robustness in safety-critical settings (\emph{Heaven}) yet risks representational harms in societal contexts (\emph{Sin}).}
\label{fig:valuation_paradox}

\end{figure}
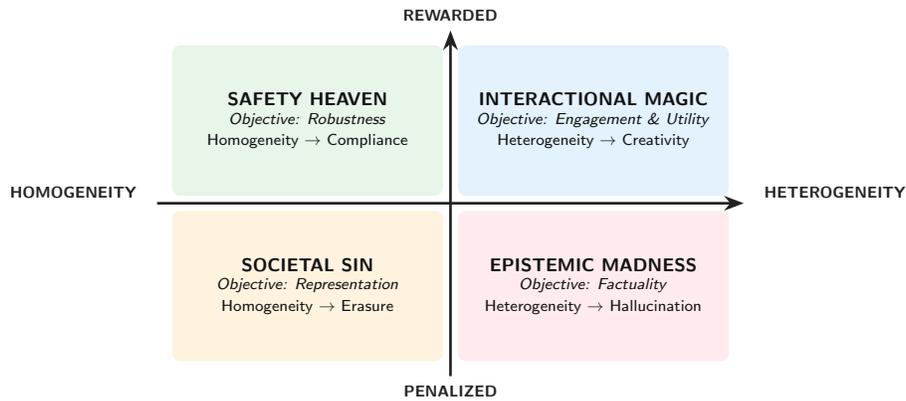


This observation gives rise to a simple but powerful structure defined by two dimensions: the degree of output variation and its normative valuation, that is, whether the observed variation is rewarded or penalized. Together, these define a conceptual space (Figure \ref{fig:valuation_paradox}) within which we identify the following four normative contexts:

\begin{itemize}
    \item \textbf{Epistemic (Section \ref{sec:epistemic}):} The objective is factual correctness and reasoning reliability. Heterogeneity is penalized as hallucination or error. Hence, \emph{madness}.
    
    \item \textbf{Interactional (Section \ref{sec:interactional}):} The objective is user utility and engagement. Heterogeneity is valued as creativity and exploration. Hence, \emph{magic}.
    
    \item \textbf{Societal (Section \ref{sec:societal_sin}):} The objective is fair representation. Homogeneity is penalized as erasure and stereotyping. Hence, \emph{sin}.
    
    \item \textbf{Safety (Section \ref{sec:safety_heaven}):} The objective is robustness. Homogeneity is valued as robust alignment and compliance. Hence, \emph{heaven}.
\end{itemize}

We refer to this as the \textbf{Magic, Madness, Heaven, Sin} framework, after the four quadrants. 

It should be noted that these contexts are not exhaustive, but represent the dominant normative lenses through which output variation is currently evaluated in the literature \cite{huang2025trustworthinessgenerativefoundationmodels, kashyap-etal-2025-helpful, liu2024trustworthyllmssurveyguideline}. Also note that individual tasks (like, recommendations in the interactional context) may not fully align with their context's dominant valuation, or tasks may operate at different levels of analysis where variation is valued differently. We explicitly note such cases throughout the paper.

\begin{figure*}[t]
\centering
\resizebox{\textwidth}{!}{%
\begin{tikzpicture}[
    querybar/.style={draw=black!60, line width=0.8pt, fill=white, rounded corners=4pt, inner sep=6pt, text width=15cm, align=center, drop shadow={opacity=0.05, shadow xshift=0pt, shadow yshift=-1pt}},
    contextbox/.style={draw=black!50, line width=0.8pt, rounded corners=4pt, inner sep=8pt, text width=4.8cm, minimum height=2.4cm, align=center, drop shadow={opacity=0.05, shadow xshift=0pt, shadow yshift=-1pt}}
]

\definecolor{magicColor}{HTML}{E3F2FD} 
\definecolor{madnessColor}{HTML}{FFEBEE} 
\definecolor{heavenColor}{HTML}{E8F5E9} 

\node[querybar] (query) at (0,0) {
    \textbf{User Query:} \textit{``How should I manage my severe chronic pain?''}
};


\node[contextbox, fill=heavenColor] (safety) at (-5.3, -1.6) {
    \textbf{Safety Context}\\
    \vspace{0.05cm}
    \footnotesize\textit{Objective: Robustness}\\
    \vspace{0.15cm}
    \small \textbf{Homogeneity}\\
    \scriptsize (Consistently filter out harmful advice or dangerous dosages)
};

\node[contextbox, fill=madnessColor] (epistemic) at (0, -1.6) {
    \textbf{Epistemic Context}\\
    \vspace{0.05cm}
    \footnotesize\textit{Objective: Factuality}\\
    \vspace{0.15cm}
    \small \textbf{Homogeneity}\\
    \scriptsize (Strictly adhere to verified medical facts and drug interactions)
};

\node[contextbox, fill=magicColor] (interactional) at (5.3, -1.6) {
    \textbf{Interactional Context}\\
    \vspace{0.05cm}
    \footnotesize\textit{Objective: Utility}\\
    \vspace{0.15cm}
    \small \textbf{Heterogeneity}\\
    \scriptsize (Provide a diverse range of treatments like medicines, therapy, etc.)
};

\end{tikzpicture}%
}
\definecolor{safetyred}{RGB}{180, 60, 60}
\definecolor{epistemicblue}{RGB}{70, 100, 170}
\definecolor{intgreen}{RGB}{0, 128, 80}
\caption{\textbf{Application of Magic, Madness, Heaven, Sin Framework.} Applying the framework to the query reveals three active objectives with competing valuations. The \textcolor{epistemicblue}{epistemic objective (factuality)} demands \emph{homogeneity} — the model should converge on verified medical facts. The \textcolor{safetyred}{safety objective (robustness)} demands \emph{homogeneity} — the model should consistently avoid dangerous dosages or unverified treatments. The \textcolor{intgreen}{interactional objective (utility)} demands \emph{heterogeneity} — the model should surface a diverse range of treatment options. The ideal response must converge on safe, factually grounded content while diverging in the space of options presented.}
\label{fig:query_tension}
\end{figure*}
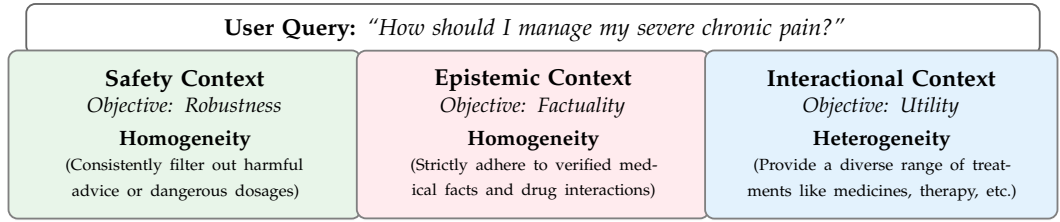

In practice, a single task often activates multiple normative objectives simultaneously, imposing competing demands on output variation (Figure \ref{fig:query_tension}). We analyze these tensions in Section \ref{sec:discussion}.

Our contributions are as follows:

\textbf{1. Unified framework for LLM output variation.} We introduce a framework that organizes LLM output variation along a homogeneity-heterogeneity axis, where the valuation of variation is determined by the task and its normative objective. This reframes variation as a context-dependent property rather than a model's intrinsic trait.

\textbf{2. Cross-contextual vocabulary mapping.} We instantiate this framework through four normative contexts — epistemic, interactional, societal, and safety — and for each context, examine the tasks, failure modes, and terminology through which output variation is studied, providing a shared vocabulary across otherwise siloed research areas. We also apply our framework to determine the valuation of variation for each task.

\textbf{3. Application \& Cross-contextual analysis.} We show how our framework can be applied to any new setting — by identifying active normative objectives and their valuations — and demonstrate that at the system level, optimizing for one objective can structurally degrade another. Through a systematic pairwise analysis of all six cross-contextual interactions, we reveal structural tensions that arise when optimizing across competing normative objectives.


\section{Epistemic Context}
\label{sec:epistemic}

In the Epistemic Context, the normative objective is factual correctness and reasoning reliability. User queries that seek factual answers or require structured reasoning assume that there exists a correct answer (or a small, well-defined set of correct answers), and the model is expected to reliably converge to it. Here, output heterogeneity is the primary failure mode: divergence from ground truth manifests as hallucination or logical inconsistency. We examine these failures in two settings: fact-based question answering and reasoning-intensive tasks.

\subsection{Task: Fact-based QA}

User queries that seek factual answers require LLMs to produce responses that are accurate and grounded in real-world knowledge \cite{wang2023surveyfactualitylargelanguage, pan2025llmsrefusequestionsknow}. For example, a question like \textit{``What is the capital of France?''} admits a canonical answer, and any deviation from it would be considered as an error. In this setting, heterogeneity constitutes epistemic failure, commonly referred to as \textbf{hallucination}: the model generates content that is plausible-sounding but factually incorrect or unsupported \cite{Ji_2023, Huang_2025, Kim2025.02.28.25323115}.

Beyond outright factual errors, heterogeneity also manifests as miscalibrated confidence. Language models frequently exhibit overconfidence, producing incorrect answers with high certainty, which can mislead users. To address this, \textbf{uncertainty calibration} \cite{kadavath2022languagemodelsmostlyknow, liu2025uncertaintyquantificationconfidencecalibration} and \textbf{abstention} become important. Ideally, when the model lacks sufficient knowledge, it should either express uncertainty or refrain from answering altogether \cite{yin2023largelanguagemodelsknow, pan2025llmsrefusequestionsknow}.

\subsection{Reasoning Tasks}

Structured reasoning-intensive tasks such as mathematical problem solving, logical inference, and code generation typically involve the model constructing a sequence of intermediate steps that collectively lead to a valid solution \cite{cobbe2021trainingverifierssolvemath, uesato2022solvingmathwordproblems, wei2023chainofthoughtpromptingelicitsreasoning, abdollahi2025demystifyingerrorsllmreasoning, song2026largelanguagemodelreasoning}.

In this context, epistemic correctness is evaluated across two distinct dimensions: the intermediate logical trajectory and the final functional outcome. Correctness is sensitive to the entire reasoning trajectory \cite{liu2025enhancingmathematicalreasoninglarge, gao2025systematicliteraturereviewcode, abdollahi2025demystifyingerrorsllmreasoning}. Heterogeneity manifests as variability in reasoning trajectories, where the model may produce multiple distinct chains of thought, some of which lead to logically inconsistent and incorrect results. Reliable behavior therefore avoids divergence in the underlying reasoning process.

However, the final outcome exhibits a different relationship with variation. Epistemic tasks often tolerate heterogeneity at the surface level: lexical, structural, or syntactic, provided the semantic or functional output stays consistent. For example, a model may generate syntactically diverse code \cite{gao2025systematicliteraturereviewcode, shypula2025evaluatingdiversityqualityllm} as long as all variants execute to the same correct result.

This illustrates that even within a single context, the valuation of variation can differ across levels of analysis: epistemic tasks penalize semantic heterogeneity while remaining agnostic to surface-level variation.

\section{Interactional Context}
\label{sec:interactional}

In the Interactional Context, the normative objective is user utility, engagement, and novelty. Most tasks such as creative writing, brainstorming, and open-ended QA reward heterogeneity, with some exceptions where personalization favors a degree of homogeneity. We examine this across three settings: creative writing, open-ended dialogue, and open-ended question answering.

\subsection{Task: Creative Writing and Brainstorming}
\label{subsec:creative_writing}

Users often employ LLMs for tasks that require exploring the long tail of the output distribution, producing responses that are novel, surprising, and meaningfully distinct. Such expectations commonly arise in research brainstorming \cite{liao2024llmsresearchtoolslarge}, creative writing \cite{MOON2025100207}, story generation \cite{doi:10.1073/pnas.2504966122}, and open-ended ideation \cite{10.1145/3613904.3642414}. In these settings, output heterogeneity is the normative objective: users value divergence across and within responses. Convergence to predictable or repetitive outputs constitutes the primary failure mode, commonly referred to as \textbf{homogenization}.

Recent work documents this failure at multiple levels. At the \emph{semantic level}, \cite{MOON2025100207} show that the marginal diversity contributed by each additional model-generated essay decreases as the corpus grows — more rapidly than for human-written essays. \cite{jiang2025artificialhivemindopenendedhomogeneity} find that LLMs exhibit pronounced \textbf{mode collapse} \cite{zhang2025verbalizedsamplingmitigatemode}, converging to a narrow output space they term the \textit{Artificial Hivemind}. This collapse operates along two dimensions: \textit{intra-model}, where repeated samples converge to the same ideas, and \textit{inter-model}, where independently trained models produce similar responses. At the \emph{narrative level}, \citet{doi:10.1073/pnas.2504966122} further show that even when LLMs generate lexically distinct stories, the underlying plot elements remain highly redundant.


Interactional settings also value heterogeneity in \emph{how} ideas are expressed, not just \emph{what} ideas are generated. Users value the texture of human writing, where unpredictable shifts in tone and rhythm prevent monotony. Researchers quantify this through \textbf{Perplexity} (token-level surprise) and \textbf{Burstiness} (clustered patterns in language use across a given text) \cite{Jelinek1977PerplexityaMO, Church1995PoissonM, tian2023identifying, info:doi/10.2196/51229}. Low scores are frequently associated with recognizably AI-like, stylistically smooth but structurally repetitive text \cite{HADAN2024100095, info:doi/10.2196/62779}.


Homogenization also extends into \emph{human-AI collaboration}. Using LLMs as creativity support tools leads different users to generate more similar ideas \cite{10.1145/3635636.3656204}, and essays co-written with aligned models exhibit lower semantic diversity than those written with base models \cite{padmakumar2024doeswritinglanguagemodels}. 

Taken together, these results point to homogenization in creative generation at multiple levels: within individual models, across models, and in human-AI co-written outputs.

\subsection{Task: Open-Ended Dialogue}
\label{subsec:dialogue}

Open-ended dialogue involves multi-turn interaction in which output variation operates at two levels. Across users, heterogeneity is desirable: different users should receive distinct conversational trajectories conditioned on their intent, preferences, and interaction history. Within a single user's interaction, however, homogeneity is expected: responses should remain consistent with that user's established preferences. This tension arises from \textbf{personalization} requirements to maintain user engagement and cater to their utility \cite{wan2025enhancingpersonalizedmultiturndialogue, wang2025enhancinguserengagementsociallydriven}.

Current optimization frameworks struggle with this balance. By averaging over diverse human preferences, they produce homogenized policies that fail to differentiate across users \cite{10.5555/3737916.3739580, wang-etal-2024-learning-personalized, yunusov-etal-2025-personality} — a limitation that has motivated work on \textbf{pluralistic alignment} \cite{sorensen2024roadmappluralisticalignment}. At the other extreme, models can over-align with a user's expressed beliefs or framing through \textbf{sycophancy}, collapsing into simple agreement rather than offering genuinely personalized responses \cite{cheng2025elephantmeasuringunderstandingsocial, sharma2025understandingsycophancylanguagemodels, hong-etal-2025-measuring}.



\subsection{Task: Open-Ended QA}
\label{subsec:open_ended_qa}

Open-ended user queries do not resolve to a single canonical response, but instead span a set of reasonable alternatives \cite{jiang2025artificialhivemindopenendedhomogeneity}, making output heterogeneity essential for capturing this breadth. This heterogeneity can be realized in different ways: models may preserve diversity across multiple responses, offering distinct answers upon repeated sampling (\emph{inter-response diversity}), or compress that diversity within a single response by aggregating multiple perspectives (\emph{intra-response pluralism}). Current models tend to favor consolidated, internally pluralistic answers at the expense of variability across responses \cite{sorensen2024roadmappluralisticalignment, lake2025distributionalovertonpluralisminvestigating}. We now turn to the types of open-ended questions commonly encountered in real-world user interactions \cite{jiang2025artificialhivemindopenendedhomogeneity}.

\paragraph{Subjective and Multi-perspective QA.}
This class of questions is common in open-ended contexts where the answer space inherently contains multiple valid viewpoints rather than a single agreed-upon response. Prior work characterizes such questions as involving either opposing, binary stances on subjective claims \cite{chen-etal-2019-seeing}, or broader multi-faceted information needs with \textit{unknown unknowns} where no single perspective is sufficient for completeness \cite{rosset2024researchyquestionsdatasetmultiperspective}. Here, heterogeneity reflects \textbf{coverage} over different, plausible valid perspectives, stances, or viewpoints that have grounded evidence for support \cite{chen-etal-2019-seeing, hayati-etal-2024-far, lv-etal-2024-subjective}.

\paragraph{Ambiguous QA.}
Ambiguity is inherent to open-domain QA, where a query's surface form often supports multiple plausible interpretations due to latent underspecification of user intent \cite{min-etal-2020-ambigqa, ji2025deepambigqaambiguousmultihopquestions}. For example, \textit{"Who was the President of the United States in 2025?"} is temporally ambiguous: depending on whether the reference point is before or after January 20, both Joe Biden and Donald Trump are valid answers. When models confidently converge on a single interpretation, they produce incomplete responses that may not address the user's intended query \cite{shi-etal-2025-ambiguity}, necessitating heterogeneity in the answer space to ensure \textbf{completeness} \cite{min-etal-2020-ambigqa, 10.1145/3471158.3472257, aliannejadi-etal-2021-building, 10.1145/3746059.3747686}.


\paragraph{Recommendations.} Unlike other interactional QA tasks, the primary driver for recommendations is \textbf{personalization}, which favors homogeneous outputs aligned with user preferences \cite{kantharuban-etal-2025-stereotype, neplenbroek-etal-2025-reading}. However, excessive alignment can narrow the recommendation space, giving rise to \textbf{filter bubble} effects \cite{areeb2023filterbubblesrecommendersystems}. Hence, recommendations must balance heterogeneity by surfacing novel options with the homogeneity driven by personalization.
\section{Societal Context}
\label{sec:societal_sin}

In the Societal Context, the normative objective is fair representation. Here, homogeneity is the failure mode: when models converge on dominant demographic, cultural, or ideological defaults, they erase the heterogeneity of human populations. We examine this across three dimensions: demographic representation, cultural representation, and values.

\subsection{Demographic Representation}
\label{subsec:demographics}

Extensive research has documented the tendency of LLMs to propagate \textbf{social biases} \cite{gallegos-etal-2024-bias} against protected demographic groups, including gender, sexual orientation, age, disability, nationality, and race \cite{sheng-etal-2021-societal, dhingra2023queerpeoplepeoplefirst, hassan-etal-2021-unpacking-interdependent, 10.1145/3582269.3615599, Yang2024RacialBiasMedReports, 10.1145/3663547.3746464, pelosio2025obscurederasedevaluatingnationality}. 
Within the societal context, this bias manifests as a systemic tendency toward homogeneity, which we characterize through two mechanisms.

The first is \textbf{erasure}: certain demographic groups are rendered statistically invisible across model generations. With underspecified prompts, models disproportionately converge on majority populations, effectively erasing minority identities from generated content \cite{lahoti-etal-2023-improving, cheng-etal-2023-marked, dhingra2023queerpeoplepeoplefirst}. This is particularly severe for non-binary \cite{dev-etal-2021-harms} and transgender communities \cite{blodgett2021justnlp}, whose identities are often absent from both training data and model outputs.


The second is \textbf{stereotyping}: the propagation of reductive generalizations about particular social groups \cite{blodgett-etal-2020-language}. When demographic identities are represented in model outputs, they are systematically constrained to rigid identity-role associations, disproportionately affiliating marginalized identities with stigmatized contexts \cite{sheng-etal-2021-societal, 10.1145/3582269.3615599}. Recent work shows that text generated about minority identities exhibits higher determinism, which homogenizes their narratives and reduces the representational complexity of their lived experiences \cite{lee2025tokensamplinguncertaintydoes}.

Collectively, these mechanisms constitute \textbf{representational harms} \cite{Barocas2018FairnessAM, blodgett-etal-2020-language}: models either deny the existence of marginalized groups (erasure) or distort their lived reality (stereotyping). These distortions can serve as precursors to \textbf{allocational harms}, where biased representations propagate into downstream decision-making systems \cite{blodgett-etal-2020-language}. In societal contexts, increasing heterogeneity in model outputs is therefore the normative objective \cite{gallegos-etal-2024-bias}.

\subsection{Cultural Representation}
\label{subsec:culture}

In NLP research, culture is rarely explicitly defined. Instead, it is typically operationalized through proxies such as geographical region and language, or semantic domains like food, political relations, social etiquette, and cultural values \cite{adilazuarda-etal-2024-towards}. Within the cultural context, \textbf{social bias} manifests as a systemic tendency toward homogeneity, that is, a convergence toward Anglocentric identities and values, stereotypical associations linked to specific nationalities, and reduced representational complexity for non-Western cultures \cite{Wdowicz2025Caricature, pelosio2025obscurederasedevaluatingnationality,qadri2025risksculturalerasurelarge}. 

Our focus here is on cultural knowledge and commonsense — the extent to which models possess and retrieve information about languages, dialects, social norms, and geopolitical contexts. We defer discussion of the values (cultural and otherwise) models implicitly encode to the following subsection.

Heterogeneity in cultural knowledge is the normative objective as models should represent diverse regions, languages, and cultural identities. In practice, however, LLMs exhibit a skewed distribution, performing significantly better on questions about Western countries(like the United States) than non-Western regions \cite{shen-etal-2024-understanding, naous-etal-2024-beer, alkhamissi-etal-2024-investigating, cao-etal-2023-assessing}. LLMs also yield culturally different answers to the same question depending on the language it is asked in \cite{cao-etal-2023-assessing, shen-etal-2024-understanding}. Furthermore, low-resource languages and dialectal varieties are systematically underrepresented in training corpora \cite{khanna2025invisiblelanguagesllmuniverse, nguyen-etal-2024-democratizing, pan-etal-2025-analyzing}, reducing the cultural diversity encoded in model outputs.


\subsection{Values and Politics}
\label{subsec:values}

Beyond knowledge and commonsense, LLMs also encode cultural, moral, political, and socio-economic value systems. Prior work suggests that these values are homogenized in line with WEIRD societies (Western, Educated, Industrialized, Rich, Democratic) \cite{zhou2025llmsweirdexploringweirdness}. Multiple studies show that default-mode LLMs tend to align morally and culturally with United States norms and English-speaking Protestant European countries \cite{johnson2022ghostmachineamericanaccent, Tao_2024, benkler2023assessingllmsmoralvalue}. Political analyses similarly demonstrate systematic ideological leanings \cite{feng-etal-2023-pretraining}. This homogenization arises in part from skewed training data distributions and uneven global internet participation \cite{zhou2025llmsweirdexploringweirdness, ali2025operationalizingpluralisticvalueslarge}.




\section{Safety Context}
\label{sec:safety_heaven}

In the Safety Context, the normative objective is robustness: models must consistently adhere to prescribed behavioral constraints regardless of how they are prompted \cite{liu2025scalesjustitiacomprehensivesurvey, hui2025tridentbenchmarkingllmsafety}. Here, homogeneity is valued, and any deviation from safe behavior is treated as a failure mode. These objectives are operationalized through alignment techniques such as supervised fine-tuning and preference optimization \cite{christiano2023deepreinforcementlearninghuman, ouyang2022traininglanguagemodelsfollow, bai2022constitutionalaiharmlessnessai, rafailov2024directpreferenceoptimizationlanguage, yuan2025hardrefusalssafecompletionsoutputcentric}, and at deployment time through guardrails that screen inputs and outputs \cite{dong2024buildingguardrailslargelanguage}. We categorize these safety behaviors into \textbf{refusal} and \textbf{compliance}: refusal enforces \emph{negative constraints} by excluding unsafe regions of the output space, whereas compliance enforces \emph{positive constraints} by requiring adherence to specific external standards which we will discuss more below.

\subsection{Refusal and Safe Completion}

A core safety objective in LLMs is the \textit{refusal} of policy-prohibited content, including instructions for weapons, illicit substances, malware, hate speech, copyright reproduction, and private data disclosure \cite{wang-etal-2024-answer, yuan2025hardrefusalssafecompletionsoutputcentric}. More importantly, this refusal must be robust under adversarial perturbation; models are expected to maintain consistent safety behavior despite paraphrasing, role-play framing, translation, jailbreaking, or prompt injection attempts \cite{liu2025scalesjustitiacomprehensivesurvey}.

Recent work distinguishes \textbf{safe completion} as a softer alternative to rigid refusal. In this regime, the model avoids providing actionable harmful details while still addressing the user's underlying intent in a helpful manner \cite{yuan2025hardrefusalssafecompletionsoutputcentric}.

In this setting, the constraints enforce safety by steering the model away from outputs it should not produce.

\subsection{Compliance}

In high-stakes fields such as medicine, law, and finance, the consequences of error are high; a single incorrect response can result in substantial liability, financial loss, or physical harm, including loss of life. Examples of unsafe model behavior include providing unethical financial guidance, suggesting illegal actions, or proposing unverified medical guidance \cite{hui2025tridentbenchmarkingllmsafety}. Because these domains are highly regulated, safety is defined by consistent adherence to established professional, legal, and ethical standards that govern acceptable practice \cite{Mesko2023, Kelsall2025, o'neill2026a}.

In enterprise and commercial LLMs, safety manifests as a strict requirement for \textit{determinism} and \textit{auditability}. Businesses usually require models to produce reproducible, brand-aligned responses, ensuring that the customer experience remains consistent rather than dependent on stochastic generation \cite{prabhune2025informationconsistentlanguagemodelrecommendations}. For example, if a commercial chatbot provides inconsistent answers to the same question, it is treated as a compliance failure. Furthermore, several legal frameworks mandate traceability and auditability in high-stakes applications to ensure safety and compliance \cite{hui2025tridentbenchmarkingllmsafety}.

In this setting, the constraints enforce safety by steering the model toward outputs that it is expected to produce.
\section{Discussion}
\label{sec:discussion}

In Sections \ref{sec:epistemic} through \ref{sec:safety_heaven}, we examined how model output variation is studied within each normative context in isolation, establishing a shared vocabulary across tasks and domains.

\subsection{Applications of framework}

In the preceding sections, we applied the framework at the task level: for each task, we identified the dominant normative objective and determined whether output variation is rewarded or penalized. The same reasoning extends to any new setting: given a task, one first identifies the active normative objectives, then determines the valuation of output variation conditioned on the task and each objective. For example, the query ``How should I manage my severe chronic pain?'' activates epistemic, safety, and interactional objectives — the first two demanding homogeneity, the third demanding heterogeneity (Figure \ref{fig:query_tension}). The ideal response must satisfy these competing demands simultaneously.

Beyond individual tasks, the framework applies at the system level. Training interventions that optimize for one objective reshape the model's entire output distribution, with consequences across all contexts. Table~\ref{tab:system_tradeoffs} maps these tensions across all six pairwise interactions, organized into three categories: opposing behaviors on the axis, same behavior with different valuations, and same behavior with same valuation but different normative context.



\begin{table*}[t]
\centering
\small
\renewcommand{\arraystretch}{1.4}
\begin{tabular}{@{} p{0.48\linewidth} p{0.48\linewidth} @{}}
\toprule

\multicolumn{2}{@{}l}{\textbf{Category 1: Opposing behaviors (X-axis)}} \\
\midrule
\textbf{Safety $\times$ Interactional} & \textbf{Societal $\times$ Epistemic} \\
Safety requires output convergence, while interactional settings reward divergence for creativity. The homogenization effects documented in Section \ref{subsec:creative_writing} are widely attributed to alignment and post-training procedures driven by safety objectives \cite{kirk2024understandingeffectsrlhfllm, MOON2025100207, padmakumar2024doeswritinglanguagemodels}.
& 
Societal goals penalize convergence as erasure, while epistemic goals penalize divergence as inconsistency. Optimizing for factual consistency can suppress culturally relevant variation \cite{cao-etal-2023-assessing, shen-etal-2024-understanding}. Moreover, in medical contexts, omitting protected attributes like age and race can avoid stereotypes but can degrade diagnostic accuracy, as these features are often clinically relevant \cite{aghaebe-etal-2025-llms}. \\

\midrule
\multicolumn{2}{@{}l}{\textbf{Category 2: Same behavior, different valuations (Y-axis)}} \\
\midrule
\textbf{Safety $\times$ Societal} & \textbf{Interactional $\times$ Epistemic} \\
Safety rewards output homogeneity as robust alignment, while societal context penalizes homogeneity as representational erasure. Alignment techniques that improve safety can compress demographic \& cultural diversity \cite{ali2025operationalizingpluralisticvalueslarge}. Also, reward models trained for safety penalize non-standard dialects such as African American English \cite{mire-etal-2025-rejected}, reinforcing linguistic erasure.
& 
Interactional contexts reward output heterogeneity as creative novelty, while epistemic context penalizes heterogeneity as factual unreliability. First, Sycophancy, a failure mode of Interactional context, can lead models to abandon correct answers across multi-turn dialogue \cite{cheng2025elephantmeasuringunderstandingsocial, sharma2025understandingsycophancylanguagemodels}. Second, methods that suppress uncertainty to improve factuality can reduce expressive diversity, leading to less creative writing \cite{sui2026llmsexhibitsignificantlylower}. \\

\midrule
\multicolumn{2}{@{}l}{\textbf{Category 3: Same behavior, same valuation, different context (Normative Context)}} \\
\midrule
\textbf{Societal $\times$ Interactional} & \textbf{Safety $\times$ Epistemic} \\
Both contexts favor heterogeneity and penalize homogeneity, but of qualitatively different kinds. Societal heterogeneity demands distributional coverage across demographic groups, while interactional heterogeneity demands novelty and user-specific adaptation. One caveat is Personalization (Section \ref{subsec:dialogue}): as it promotes homogeneity for individual users, it can reinforce stereotypical associations \cite{kantharuban-etal-2025-stereotype, neplenbroek-etal-2025-reading}.

& 
Both contexts favor homogeneity. Safety-driven behavioral convergence reinforces epistemic convergence toward factual correctness, and this co-optimization is reflected in standard alignment pipelines \cite{askell2021generallanguageassistantlaboratory}. This explains why societal and interactional goals, which require heterogeneity, are systematically disadvantaged by the same process \cite{kirk2024understandingeffectsrlhfllm, Murthy_2025}. \\

\bottomrule
\end{tabular}
\caption{\textbf{System-level tradeoffs between normative contexts.} Reading top-to-bottom demonstrates the progressive complexity of alignment tensions: from directional conflicts on the X-axis (Category 1), to valuation conflicts on the Y-axis (Category 2), to substantive conflicts driven by the normative contexts themselves (Category 3).}
\label{tab:system_tradeoffs}
\end{table*}


\subsection{Contemporary Frameworks and Scope}

Recent work has proposed frameworks for measuring output diversity across lexical, syntactic, and semantic dimensions \cite{guo2025benchmarkinglinguisticdiversitylarge}, and for disentangling diversity from quality through metrics like effective semantic diversity \cite{shypula2025evaluatingdiversityqualityllm}. Our framework abstracts above this measurement layer — agnostic to the specific level of analysis (lexical, semantic, inter-model, intra-response, etc.) or metric employed — providing a complementary normative lens that determines whether observed variation should be rewarded or penalized given the task and its objective.

A prominent evaluative framework for LLM behavior is HHH(Helpful, Harmless, Honest) \cite{askell2021generallanguageassistantlaboratory}, which defines three alignment criteria and acknowledges conflicts between them. Our framework generalizes to any normative objective, including societal ones which HHH does not directly address, and grounds all dimensions in a shared axis of output variation, making cross-contextual tensions structurally explicit. Independent concurrent work by \citet{riossialer2026structureawarediversitypursuitai} and \citet{esteve2026surveydiversityquantificationnatural} similarly argue that diversity should be evaluated relative to context, with the latter advocating for a normative lens. Our framework provides such a formalization, grounded in task objectives.


Some limitations should be noted. The four contexts are demonstrative rather than exhaustive. The framework abstracts away from mechanisms that modulate variation (sampling strategies, training stages, prompting techniques), focusing on valuation rather than origins. Disentangling levels of analysis and metrics is beyond our scope (see Table \ref{tab:diversity_taxonomy} in Appendix). Finally, while the framework reveals conflicts between objectives, it does not prescribe resolutions. Developing context-aware control mechanisms remains future work.

\section{Conclusion}

We introduced the Magic, Madness, Heaven, Sin framework, which models LLM output variation along a continuous homogeneity–heterogeneity axis, where the valuation of variation is determined by the task and its normative objective. By organizing tasks into four normative contexts — epistemic, interactional, societal, and safety — we showed that variation is not a model's intrinsic trait but a context-dependent property: the same output behavior can be rewarded as creativity or penalized as hallucination, valued as robust compliance or criticized as representational erasure. Through a systematic pairwise analysis of all six cross-contextual interactions, we demonstrated that optimizing for one normative objective can structurally degrade another. We hope this framework provides a shared vocabulary for researchers across NLG, alignment, fairness, and safety to reason about output variation in a more principled and context-aware manner.

\clearpage


\section*{Acknowledgments}
The author of this paper would like to thank Emily Sheng for their feedback on an earlier draft, which helped refine the framing and strengthen the overall argument.



\bibliography{custom}
\bibliographystyle{colm2026_conference}

\appendix
\clearpage
\section{Disambiguating Diversity in Post-Training}

\begin{table}[h] 
\centering
\footnotesize 
\renewcommand{\arraystretch}{1.4} 
\begin{tabularx}{\textwidth}{
  >{\raggedright\arraybackslash}p{1.5cm} 
  >{\raggedright\arraybackslash}p{2.4cm} 
  >{\raggedright\arraybackslash}p{2.4cm} 
  >{\raggedright\arraybackslash}p{1.8cm} 
  >{\raggedright\arraybackslash}X 
}
\toprule
\textbf{Study} & \textbf{Level of Diversity} & \textbf{Type of Diversity} & \textbf{ML Task} & \textbf{Key Conclusion (Alignment $\leftrightarrow$ Diversity)} \\
\midrule

\citet{kirk2024understandingeffectsrlhfllm} & 
Lexical \& Semantic & 
Inter-response (per-input, across inputs) & 
Summarization & 
RLHF helps increase model's out-of-distribution generalization capability at the expense of \textbf{loss of output diversity} \\

\citet{guo2025benchmarkinglinguisticdiversitylarge} & 
Lexical, Syntactic \& Semantic & 
Inter-Response (across inputs) & 
5 NLG Tasks \newline (like Summarization \& Story Generation) & 
Instruction tuning tends to \textbf{increase lexical diversity} while \textbf{reducing syntactic and semantic diversity} compared to base variants \\

\citet{shypula2025evaluatingdiversityqualityllm} & 
Lexical, Syntactic, Semantic \& Effective Semantic (output quality controlled) & 
Inter-response (per-input) & 
Open-ended Code Generation & 
Preference-tuned models exhibit reduced lexical and syntactic diversity but produce \textbf{greater effective semantic diversity} than SFT or base models, primarily because of more valid outputs \\

\citet{lake2025distributionalovertonpluralisminvestigating} & 
Lexical, Semantic & 
Intra-response, Inter-response (per-input) & 
Open-ended QA & 
Alignment reduces inter-response diversity but \textbf{increases intra-response pluralism} by aggregating multiple perspectives into single, longer responses \\

\citet{sorensen2025spectrumtuningposttrainingdistributional} & 
Coverage over valid outputs \& Semantic & 
Inter-Response (per-input) & 
Verifiable Generation Tasks (with many valid answers), Open-ended chat completion & 
Post-trained (instruction-tuned) models exhibit \textbf{reduced diversity}, characterized by lower coverage of valid outputs and reduced variability across responses. \\

\citet{Murthy_2025} & 
Conceptual & 
Inter-response (per-persona, across personas) & 
Association; Similarity Judgments & 
RLHF/RLAIF \textbf{reduces diversity} in the associations and perspectives expressed across simulated personas, relative to instruction fine-tuned models \\

\bottomrule
\end{tabularx}
\caption{\textbf{Disambiguating Diversity in Post-Training.} A comparison of key studies reveals that ``diversity'' refers to distinct phenomena: from lexical variation to conceptual coverage across an array of different ML tasks.}
\label{tab:diversity_taxonomy}
\end{table}

\end{document}